\documentclass{article}

\usepackage[pdftex]{pstricks}
\usepackage{verbatim}
\usepackage{amsmath}
\usepackage{subfigure}
\usepackage{graphicx}
\usepackage[linesnumbered,vlined,ruled]{algorithm2e}

\usepackage[french]{babel} 
\usepackage[utf8]{inputenc} 
\usepackage{lmodern} 
\usepackage{xspace} 
\AddThinSpaceBeforeFootnotes 
\FrenchFootnotes 

\title{MGP: Un algorithme de planification temps réel prenant en compte l'évolution dynamique du but}

\author{Damien Pellier$^{1}$, Mickaël Vanneufville$^{1,2}$,  Humbert Fiorino$^{2}$, \\ Marc Métivier$^{1}$ and Bruno Bouzy$^{1}$ \\[4mm]
$^{1}$Laboratoire d'Informatique de Paris Descartes \\
45, rue des Saints Pres, 75006 Paris\\[4mm]
$^{2}$Laboratoire d'Informatique de Grenoble\\
110 avenue de la Chimie 38400 Saint-Martin-d'Hères
}
\date{Juin 2012}

\begin{document}
\maketitle

\begin{abstract}
  Dans cet article, nous proposons un nouvel algorithme de planification temps réel appelé MGP ({\it Moving Goal Planning}) capable de s'adapter lorsque le but évolue dynamiquement au cours du temps. Cet algorithme s'inspire des algorithmes des type {\it Moving Target Search} (MTS). Afin de réduire le nombre de recherches effectuées et améliorer ses performances, MGP retarde autant que possible le déclenchement de nouvelles recherches lorsque que le but change. Pour cela, MGP s'appuie sur deux stratégies: {\it Open Check} (OC) qui vérifie si le nouveau but est présent dans l'arbre de recherche déjà construit lors d'une précédente recherche et {\it Plan Follow} (PF) qui estime s'il est préférable d'exécuter les actions du plan courant pour se rapprocher du nouveau but plutôt que de relancer une nouvelle recherche. En outre, MGP utilise une stratégie "conservatrice" de mise à jour incrémentale de l'arbre de recherche lui permettant de réduire le nombre d'appels à la fonction heuristique et ainsi d'accélérer la recherche d'un plan solution. Finalement, nous présentons des résultats expérimentaux qui montrent l'efficacité de notre approche.
\end{abstract}

\section{Introduction}

Dans des environnements dynamiques, les agents doivent constamment s'adapter pour faire face aux événements qui viennent perturber leurs plans en entrelaçant planification et exécution. On parle alors de planification en boucle fermée. Il n'est pas rare que ces évènements conduisent à une modification du but initial qui leur a été confié. Ce cas se produit notamment lorsqu'un agent est en interaction avec un opérateur humain qui peut venir à tout instant modifier le but précédemment assigné. La littérature distingue principalement deux manières d'aborder ce problème \cite{Nebel95planreuse, Kambhampati:1992,aaai-fs98b,Fox06planstability,Krogt05}: planifier un nouveau plan en partant de zéro ou réparer le plan courant pour prendre en compte le nouveau contexte. Bien qu'en théorie ces deux approches soient équivalentes en terme de coût dans le pire des cas \cite{Nebel95planreuse}, les résultats expérimentaux montrent que l'approche qui consiste à réparer un plan existant est plus efficace que celle qui consiste à replanifier sans réutiliser les informations provenant des précédentes recherches \cite{Krogt05}. De plus, faire évoluer aussi peu que possible les plans (stabilité des plans) est un autre argument en faveur de la premire approche \cite{Fox06planstability}.

L'adaptation aux environnements dynamiques est également un problme ouvert dans le domaine des jeux vidéo et plus particuliérement en ce qui concerne la recherche temps réel de cibles mouvantes (MTS) pour de grands espaces de recherche. Par essence, un algorithme de type MTS entrelace recherche d'un chemin et exécution: un agent {\it chasseur} poursuit une cible mouvante aussi appelée {\it proie} dans une grille à deux dimensions ou une carte. Depuis les travaux pionniers d'Ishida \cite{ishida:91,ishida:95,ishida:98}, les approches de type MTS peuvent se partager principalement en deux catégories selon la stratégie sur laquelle elles s'appuient pour réutiliser l'information des précédentes recherches.

La première stratégie consiste à utiliser une fonction heuristique pour guider la recherche et apprendre le chemin le plus court entre des couples de points sur une carte. À chaque nouvelle recherche, la fonction heuristique est plus informative et la recherche devient plus performante. Dans sa version originale, l'algorithme MTS est une adaptation de l'algorithme LRTA* ({\it Learning Real-Time A*}) \cite{korf:90}. Cette approche est complète lorsque les déplacements de la cible sont plus lents que celui de l'agent. Toutefois, elle n'est pas sans inconvénient à cause des dépressions de la fonction heuristique et à la perte d'information induite par les déplacements de la cible \cite{melax:93}. Actuellement, l'état de l'art des algorithmes fondés sur cette première stratégie sont des variantes de AA* \cite{koenig:05}, MTAA* \cite{koenig:07} et GAA* \cite{sun:08}. MTAA* adapte et apprend la fonction heuristique à partir des déplacements de la cible pour améliorer la convergence. GAA* généralise MTAA* avec la prise en compte d'actions auxquelles on a associé des coûts qui augmentent et diminuent au court du temps. MTAA* et GAA* nécessitent tout deux l'utilisation de fonctions heuristiques admissibles pour garantir leur correction et leur complétude.

La seconde stratégie consiste à utiliser incrémentalement l'arbre de recherche entre deux recherches successives. Le premier algorithme fondé sur cette stratégie est D* \cite{stenz:95}. Ses deux principaux successeurs sont décrits dans \cite{koenig:02,sun:10a}. Ces algorithmes ont été conçus et développés pour calculer rapidement des trajectoires dans des environnements inconnus et dynamiques. Ils s'appuient tous sur une recherche en chaînage arrière. Ils obtiennent de bons résultats lorsque l'environnement évolue peu au cours du temps. En revanche, ils sont surclassés par l'approche naïve qui consiste à appeler successive l'algorithme A* \cite{hart:68} à chaque fois que la cible se déplace \cite{sun:08} lorsque les seules modifications portent sur le but. Récemment, un autre algorithme s'appuyant sur une recherche en chaînage avant appelé FRA* a été proposé \cite{sun:09}. Cet algorithme n'est pas capable de prendre en compte les changements de l'environnement, mais obtient de très bons résultats quand il s'agit de prendre en compte les déplacement d'une cible au cours du temps. Chaque fois que la cible se déplace, FRA* adapte rapidement l'arbre de recherche précédemment construit à la nouvelle position de la cible et rappelle la fonction de recherche A* sur le nouvel arbre de recherche. FRA* est actuellement l'algorithme MTS le plus efficace. Toutefois, l'adaptation de l'arbre est largement dépendante de la modélisation de l'environnement. FRA* considère en effet que l'environnement est modélisé sous la forme d'une grille ce qui en limite les applications dans le cas général. Afin de pallier à cet inconvénient, une variante de cet algorithme appelée GFRA* \cite{sun:10} a été proposée pour fonctionner dans des environnements modélisés par des graphes quelconques, y compris des treillis d'états utilisé pour la navigation des véhicules terrestres autonomes. En outre, GFRA* possède une autre caractéristique intéressante: il peut être utilisé avec une fonction heuristique non admissible, ce qui est souvent le cas en planification.

Les algorithmes de recherche heuristique de type MTS ont été très peu utilisés dans le cadre de la planification. Ceci s'explique en partie par le fait qu'il a longtemps été considéré difficile de trouver des fonctions heuristiques à la fois informatives et facilement calculables dans de le cadre de la planification. Ce n'est que vers la fin des années 90 que des planificateurs tels que HSP et HSP-r \cite{bonet:hspr,bonet:hsp-aij} ont renversé la tendance en fournissant des méthodes génériques et efficaces pour calculer de telles fonctions heuristiques en temps polynomial. Ceci a eu pour conséquence de faire évoluer de manière quasiment indépendante ces deux domaines de recherche. Il nous apparaît dorénavant important de capitaliser sur les récentes avancées dans ces deux domaines pour proposer de nouveaux algorithmes de planification temps réel en boucle fermée.

Dans cet article, nous proposons un nouvel algorithme de planification temps réel appelé MGP ({\it Moving Goal Planning}) pour la planification en boucle fermée où la recherche d'un plan est vue comme une recherche heuristique de type MTS. Le reste de l'article est organisé selon le plan suivant: la partie 2 présente formellement le problème considéré et décrit l'algorithme proposé pour le résoudre; la partie 3 présente les résultats expérimentaux obtenus qui montrent l'efficacité de notre approche; finalement, la partie 4 présente nos conclusions sur le travail réalisé.

\section{Planifier avec des buts dynamiques}
\label{mgp}

Dans cette partie, nous présentons tout d'abord formellement le problème traité. Dans un deuxième temps, nous présentons l'algorithme MGP proposé et détaillons sa stratégie de recherche. Finalement, nous introduisons deux stratégies qui permettent de retarder le déclenchement d'une nouvelle recherche lorsque le but évolue dynamiquement : {\it Open Check} et {\it Path Follow}, et terminons en présentant la stratégie incrémentale de mise à jour de l'arbre de recherche.

\subsection{Modélisation du problème}

Nous nous intéressons aux problèmes de planification séquentielle de type STRIPS \cite{finke:71}. Tous les ensembles considérés sont finis. Un {\it état} $s$ est un ensemble de propositions logiques. Une {\it action} $a$ est un tuple $a = (pre(a), add(a), del(a))$ où $pre(a)$ définit les {\it préconditions} de l'action $a$ et $add(a)$ et $del(a)$ représentent respectivement ses {\it effets} positifs et négatifs. Un état $s'$ est atteignable à partir d'un état $s$ en appliquant à l'action $a$ la fonction de transition $\gamma$ suivante:
\begin{equation}
s'=\gamma(s,a) =
  \begin{cases}
    (s - del(a)) \cup add(a)  & \text{si} \ pre(a) \subseteq s\\
    \text{indéfinie} & \text{sinon}
  \end{cases}
\label{Eq:Transition}
\end{equation}
L'exécution d'une séquence d'actions  $\pi = \langle a_1, \ldots, a_n \rangle$ à une état $s$ est défini récursivement par:
\begin{equation}
\gamma(s,  \langle a_1, \ldots, a_n \rangle)= \gamma(\gamma(s, \langle a_1, \ldots, a_{n-1} \rangle), \langle a_n \rangle)
\end{equation}
Un problème de planification en boucle fermée avec but dynamique est un tuple $(A, s_t, g_t)$: à un temps donné $t$ un agent est dans un état $s_t$; $g_t$ est son but courant ($s_t$ et $g_t$ sont des ensembles de propositions) et $A$ représente l'ensemble des actions qu'il peut exécuter. L'agent exécute des actions appartenant à $A$ pour atteindre son but mais celui-ci peut évoluer dynamiquement à tout moment. L'agent ne possède aucune information sur la manière dont le but évolue au cours du temps. Cependant, nous supposons qu'à tout instant, le but $g_t$ est atteignable: un {\it plan} est une séquence d'actions $\pi = \langle a_1, \ldots, a_n \rangle$ avec $a_i \in A$ telle que $g_t \subseteq \gamma(s_t,\pi)$. $g_t$ est {\it atteignable} si un tel plan existe. Un {\it état but} est un état $s$ tel que $g_t \subseteq s$. À un temps donné $t$, un problème de planification avec but dynamique est résolu si $g_t \subseteq s_t$, autrement dit l'agent a atteint son but.

\subsection{Principe de l'algorithme}

Le pseudo-code de MGP est donné par l'algorithme \ref{Algo:MGP}. MGP prend en entrée un problème de planification $(A, s_0, g_0)$. Les variables $g$ et $s$ représentent respectivement le but courant et l'état courant initialement égaux à $g_0$ et $s_0$. $i$ est un compteur qui permet de comptabiliser le nombre de recherches effectuées.

MGP appelle itérativement une procédure de recherche (ligne~2) tant que le but courant n'a pas été atteint. La procédure de recherche {\sf Search} est détaillée dans le paragraphe \S\ref{wa}. Cette procédure construit un arbre de recherche en chaînage avant dans un espace d'états. Si le but courant n'est pas atteignable, la recherche échoue impliquant l'échec de MGP également (ligne~3). Sinon, MGP extrait un plan à partir de l'arbre de recherche construit au cours de la recherche (lignes~4-5) et essaie de retarder autant que possible un nouvel appel à la procédure de recherche, i.e., une nouvelle expansion de l'arbre de recherche (boucle {\sf While} et procédure {\sf CanDelayNewSearch} ligne~4). Cette boucle prend fin lorsque le but évolue de telle manière à ne plus être présent dans l'arbre de recherche précédemment construit. La procédure {\sf CanDelayNewSearch} est détaillée dans le paragraphe \S\ref{candelay}. Tant que le but est accessible à partir du plan précédemment extrait, MGP exécute alors les actions de ce plan (lignes~6-7). Dans le cas où, MGP atteint le but courant, MGP se termine avec succès (ligne~11). Dans le cas contraire, le but a évolué (ligne~10) et MGP doit élaguer son arbre de recherche. Son arbre de recherche est alors le sous-arbre dont la racine est représenté par son état courant $s$ ({\sf DeleteStatesOutOfTree}, line~12). Si le nouveau but est présent dans ce sous-arbre, il n'est pas nécessaire de lancer une nouvelle recherche. MGP extrait directement un nouveau plan à partir de celui-ci et exécute ses actions pour atteindre le nouveau but. Sinon, MGP met à jour les valeurs heuristiques des n{\oe}uds de l'arbre de recherche en estimant la distance pour atteindre le nouveau but (ligne~13) et finalement étend l'arbre de recherche ainsi modifié (ligne~3). La procédure de mise à jour de l'arbre de recherche {\sf UpdateSearchTree} est détaillée dans le paragraphe \S\ref{update}.

\begin{algorithm}[!]
\caption{{\sf MGP(}$A, s_0, g_0${\sf)}}
\label{Algo:MGP}
\SetKwData{False}{{FALSE}}
\SetKwData{True}{{TRUE}}
\SetKwData{Failure}{{FAILURE}}
\SetKwData{Success}{{SUCCESS}}
\SetKwData{Search}{{Search}}
\SetKwData{DeleteStatesOutOfTree}{{DeleteStatesOutOfTree}}
\SetKwData{CanDelayNewSearch}{{CanDelayNewSearch}}
\SetKwData{UpdateStatesAfterGoalChange}{{UpdateSearchTree}}
\SetKwData{UpdateGoal}{{UpdateGoal(g)}}
\SetKwData{Pathfollow}{{Pathfollow}}
\SetKwData{Open}{{open}}
\SetKwData{OpenClosed}{{open\_closed}}
\SetKwData{PF}{{PathFollow}}
\SetKwData{Null}{{NULL}}

$s \leftarrow s_0$, $g \leftarrow g_0$, $i \leftarrow 1$\;
\While{$g \not\subseteq s$}{
  \lIf{$\Search(A,s,g, i) = \Failure$}{\Return \Failure \;}
  \While{$\CanDelayNewSearch(s,g)$}{
    Extract plan $\pi = \langle a_1, \ldots, a_n \rangle$ from the search tree \;
    \While{$(g \not\subseteq s$ and  $g \subseteq \gamma(s,\pi))$ or $(\PF$ and $\pi \not= \emptyset$)}{
      Executes the first action $a_1$ in $\pi$ \;
      $s \leftarrow \gamma(s, a_1)$ \;
      $\pi \leftarrow \langle a_2, \ldots, a_n \rangle$ \;
      $g\leftarrow \UpdateGoal$\;
    }
    \lIf{$g \subseteq s$}{\Return \Success\;}
    $\DeleteStatesOutOfTree(s)$\;
  }
  $\UpdateStatesAfterGoalChange(s,g)$ \;
  $i \leftarrow i + 1$ \;
}
\end{algorithm}

\subsection{Implémentation de l'algorithme}

Dans cette partie, nous présentons les détails relatifs à l'implémentation de MGP. L'arbre de recherche est représenté par deux listes notées OPEN et CLOSED: la liste OPEN contient les états feuilles de l'arbre de recherche restants à explorer et la liste CLOSED la liste des états déjà explorés. Dans un premier temps, nous présentons la stratégie de recherche utilisée {\sf Search} (cf.~\S\ref{wa}). Puis, nous présentons la procédure {\sf CanDelayNewSearch} utilisée pour retarder les appels coûteux à la procédure de recherche. Finalement, nous présentons la procédure {\sf UpdateSearchTree} qui met à jour l'arbre de recherche entre deux recherches successives (cf.~\S\ref{update}).

\subsubsection{A* pondéré comme stratégie de recherche}
\label{wa}

Contrairement à GFRA* qui s'appuie sur une simple recherche de type A*, MGP utilise sa version pondérée. Cette variante de A* surestime volontairement la distance au but en multipliant la valeur retourné par la fonction heuristique d'un coefficient $w$. La fonction d'évaluation $f(s)$ pour un état $s$ est alors $f(s) = g(s) + w \times h(s)$ où $g(s)$ est le coût pour atteindre $s$ à partir de l'état initial $s_0$ et $h(s)$ le coût estimé pour atteindre $g$ depuis $s$. Plus $w$ est élevé, plus la fonction heuristique est prépondérante dans l'exploration de l'espace de recherche. Expérimentalement, il a été démontré que cette stratégie couplée avec une fonction heuristique informative et admissible améliore les performances en temps mais, en contrepartie, ne permet plus d'obtenir des solutions optimales \cite{pohl:70}. Toutefois, cette approche est pertinente pour le problème considéré car il est plus important de trouver rapidement une solution que de trouver un plan solution optimal. Par conséquent, utiliser une recherche pondérée de type A* peut améliorer de manière significative les performances de MGP (cf.~\S\ref{eval}). En outre, il est important de noter que l'utilisation de cette procédure de recherche n'affecte pas la correction et la complétude de MGP.

La version de A* pondéré utilisée dans notre approche (cf. Algo.~\ref{Algo:WA}) est une version légèrement modifiée de l'algorithme classique proposé par \cite{pohl:70}. Il prend en entrée un problème de planification $(A,s_0,g)$, un coefficient $w$ et le compteur $i$ qui comptabilise le nombre de recherches effectuées. À chaque état $s$, il associe quatre valeurs: sa valeur $g(s)$ et sa valeur $h(s)$, un pointeur vers son état père, $parent(s)$, dans l'arbre de recherche et l'itération à laquelle il a été créé, $iteration(s)$. Au premier appel de la procéédure, les listes OPEN et CLOSED contiennent l'état initial $s_0$ du problème de planification tel que $g(s_0) = 0$ and $h(s_0) = H(s_0, g)$ où $H$ est la fonction heuristique qui estime le coût de $s_0$ au but $g$.

La procédure de recherche A* pondérée consiste à répéter itérativement les étapes suivantes: (1) sélectionner l'état $s$ de la liste OPEN qui possède la plus petite valeur de $f(s)$ (ligne~2), (2) supprimer l'état $s$ de la liste OPEN (ligne~4), et (3) développer les états fils de $s$ en utilisant la fonction de transition $\gamma$ (cf.~ eq.~\ref{Eq:Transition}). Pour chaque état fils $s'$ obtenu avec la fonction de transition $\gamma$, si $s'$ n'est pas dans une des deux listes OPEN et CLOSED, alors A* pondéré calcule $g(s') = g(s) + c(s,s')$ où $c(s,s')$ représente le coût pour atteindre $s'$ depuis $s$. Si $s'$ n'a pas encore été exploré (lignes~8-12), $s'$ est ajouté à la liste OPEN. Sinon, A* pondéré met à jour les informations relatives à $s'$. De plus, si A* trouve un plan plus court pour atteindre $s'$ (lignes~14-17), il met alors à jour $s'$: $g(s') = g(s) + c(s,s')$ et $parent(s') = s$. Finalement, si la valeur heuristique est obsolète (lignes~18-20) parce que le but a changé, A* pondéré met à jour la valeur heuristique de $s'$.

A* pondéré possède deux conditions d'arrêt: (1) lorsque la liste OPEN est vide (aucun état contenant le but n'a pu être trouvé) et (2) lorsqu'un état contenant le but est trouvé (ligne~3). Dans ce cas, un plan solution peut être extrait de l'arbre de recherche construit par la procédure de recherche (cf.~Algo.~\ref{Algo:MGP}, ligne~5).

\begin{algorithm}[!]
\caption{{\sf Search(}$A, s, g, i${\sf)}}
\label{Algo:WA}
\SetKwData{Failure}{{FAILURE}}
\SetKwData{Success}{{SUCCESS}}
\SetKwData{True}{{TRUE}}
\SetKwData{False}{{FALSE}}

\SetKwData{Open}{{open}}
\SetKwData{Closed}{{closed}}
\SetKwData{Cost}{{cost}}

\While{$\Open \not = \emptyset$}{
  $s \leftarrow argmin_{s' \in \Open}(g(s') + w \times h(s'))$ \;
  \lIf{$g \subseteq s$}{ \Return \Success \;}
  $\Open \leftarrow \Open \setminus  \{s\}$ \;
  $\Closed \leftarrow \Closed \cup \{s\}$ \;
  \ForEach{action $a \in \{a' \in A \ | \ pre(a') \subseteq s\}$}{
    $s' \leftarrow \gamma(s, a)$\;
    $\Cost \leftarrow g(s) + c(s, s')$ \;
    \If{$s' \not \in \Open \cup \Closed$}{
      $\Open \leftarrow \Open \cup \{s'\}$ \;
      $g(s') \leftarrow \Cost$ \;
      $h(s') \leftarrow H(s', g)$ \;
      $parent(s') \leftarrow s$ \;
    } \Else{
      \If{$\Cost < g(s')$}{
        $\Open \leftarrow \Open \cup \{s'\}$ \;
        $g(s') \leftarrow \Cost$\;
        $parent(s') \leftarrow s$\;
      }
      \If{$iteration(s') < i$}{
        $h(s') \leftarrow H(s', g)$ \;
        $iteration(s') \leftarrow i$ \;
      }
    }
  }
}
\Return \Failure
\end{algorithm}

\subsubsection{Stratégies de retardement}
\label{candelay}

Afin de limiter le nombre de recherches effectuées et améliorer ses performances, MGP retarde autant que possible le déclenchement d'une nouvelle recherche lorsque le but évolue. Pour cela, MGP utilise deux stratégies originales mises en {\oe}uvre dans la procédure {\sf CanDelayNewSearch} (cf.~Algo.~\ref{Algo:MGP} ligne~4):

\paragraph{Open Check (OC)}

MGP vérifie si le nouveau but est encore présent dans l'arbre de recherche. Dans ce cas, un nouveau plan peut être extrait directement à partir de l'arbre de recherche courant. Contrairement à GFRA* qui ne teste que la liste CLOSED des états contenant les états explorés, MGP effectue également le test des états feuilles de l'arbre de recherche contenus dans la liste OPEN avant de débuter une nouvelle recherche. Cette vérification limite le nombre de recherches inutiles et les réajustements coûteux de l'arbre de recherche.

\paragraph{Plan Follow (PF)}

MGP effectue une estimation pour savoir si l'exécution de son plan courant le rapproche du but, même si celui-ci a changé. Chaque fois que le but évolue et avant de lancer une nouvelle recherche, MGP évalue si le nouveau but est proche du but précédent et détermine si le plan courant peut encore être utilisé. Cette évaluation s'appuie sur l'estimation de la fonction heuristique et le calcul d'une comparaison entre l'état courant $s$, le but précédent $p$ et le nouveau but $g$:
\begin{equation}
h(s,g) \times c > h(s,p) + h(p,g)
\label{Eq:pathfollow}
\end{equation}
$c$ est appelé le coefficient de retardement. MGP suit le plan courant tant que l'inégalité \ref{Eq:pathfollow} est vérifiée, i.e., tant qu'il estime préférable d'exécuter le plan courant pour atteindre l'ancien but puis se "diriger" vers le nouveau but plutêt que d'exécuter un plan allant directement de l'état courant $s$ au nouveau but $g$. Les valeurs de $c > 1$ permettent d'ajuster le délai avant une nouvelle recherche. étant donné que les recherches sont très coûteuses, les retarder améliore les performances de MGP mais altère la qualité des plans (cf.~\S\ref{eval}).

\subsubsection{Mise à jour incrémentale de l'arbre de recherche}
\label{update}

MGP adapte incrémentalement l'arbre de recherche à chaque nouvelle recherche (cf.~{\sf UpdateTreeSearch} Algo.~\ref{Algo:MGP} ligne~13). Contrairement à GFRA* qui met à jour la valeur heuristique de tous les états de l'arbre de recherche en estimant la distance au nouveau but, MGP s'appuie sur une stratégie moins agressive de mise à jour afin de réduire le nombre d'appels à la fonction heuristique et ainsi améliorer ses performances globales.

Dans cet objectif, MGP copie les états de la liste OPEN contenant les états feuilles de l'arbre de recherche dans la liste CLOSED, puis vide la liste OPEN  et y ajoute l'état courant en prenant soin de mettre à jour sa valeur heuristique (appel de la fonction heuristique pour estimer la distance de l'état courant au nouveau but). Pour indiquer que la valeur heuristique est à jour, MGP marque l'état avec le compteur incrémenté à chaque nouvelle recherche (cf.~Algo.~\ref{Algo:WA}, ligne~20). Au cours de la recherche, chaque fois qu'un état dans la liste CLOSED est rencontré avec une valeur d'itération inférieure à la valeur du compteur d'itération courant, l'état est ajouté dans la liste OPEN et sa valeur heuristique est mise à jour. Cette stratégie possède deux avantages: (1) elle réduit le nombre d'états créés en réutilisant les états précédemment construits et (2) réduit de manière significative le temps nécessaire à la mise à jour des valeurs heuristiques des états à partir du nouveau but en se limitant aux états explorés au cours de la nouvelle recherche.

\section{Expérimentation et évaluation}
\label{eval}

L'objectif de ces expériences est d'évaluer les performances de MGP en fonction des différentes stratégies de retardement proposées: {\it Open Check} (OC) et {\it Plan Follow} (PF) comparativement aux approches existantes. Nous avons choisi de comparer MGP à l'algorithme GFRA* qui est l'algorithme de référence dans le domaine des algorithmes de type MTS et à l'approche naïve SA* ({\it Successive A*}) qui consiste à appeler la procédure de recherche A* à chaque fois que le but évolue. Les benchmarks utilisés pour l'évaluation sont issus des compétitions internationales de planification (IPC). Nous utilisons pour l'ensemble de nos tests la fonction heuristique non-admissible proposée par \cite{hoffmann:01} dans le planificateur FF pour guider la recherche. Dans la suite, nous évaluons six algorithmes: Successive A* (SA*), Generalized Fringe-Retrieving A* (GFRA*), MGP sans stratégie de retardement (MGP), MGP avec Open Check (MGP+OC), MGP avec Plan Follow (MGP+PF) et finalement MGP avec les deux stratégies simultanément (MGP+OC+PF).

\subsection{Simulation et dynamique d'évolution du but}

Classiquement, les algorithmes MTS supposent que la cible mouvante \--- la proie \--- suit toujours le plus court chemin de sa position courante à sa nouvelle position choisie de manière aléatoire parmi les positions libres sur une grille ou une carte. Chaque fois que la proie a atteint cette position, une nouvelle position est alors sélectionnée de manière aléatoire et le processus se répète. Tous les $n$ déplacements, la proie reste immobile laissant ainsi au \--- chasseur \--- une chance de l'attraper. Cette approche n'est malheureusement pas directement transposable à la planification de tâches en boucle fermée.

Afin de l'adapter au mieux, nous avons décidé de faire évoluer le but à un pas de temps donné en lui appliquant de manière aléatoire une action parmi les actions définies dans le problème de planification. Nous faisons ici l'hypothèse que le but évolue en accord avec le modèle que l'agent a du monde. Le but que nous faisons évoluer est l'état solution obtenu par le premier appel à la procédure de recherche. En effet, dans le cas général, il peut y avoir plusieurs états but (le but d'un problème de planification de type STRIPS est un état partiel dont l'inclusion dans un état complet est requise). Le processus est répété plusieurs fois pour rendre le but plus difficile à atteindre. Notons que cette façon de faire évoluer le but garantit que le nouveau but est toujours atteignable depuis le but courant. Cependant, il n'est pas garanti que MGP soit capable d'atteindre le but puisque celui-ci peut évoluer bien plus rapidement que MGP.

Pour paramètrer l'évolution du but, un compteur $t$ est incrémenté chaque fois qu'un état est exploré au cours d'une recherche avec A* pondéré et chaque fois que la fonction heuristique est appelée pour estimer la distance au but. En effet, ces deux opérations sont de loin les plus coûteuses de MGP. À partir de ce compteur, le nombre $n$ d'actions appliquées au but est calculé comme suit:
\begin{equation}
n = (t - t_p) / g_r
\end{equation}
où $t_p$ représente la précédente valeur du compteur $t$ et $g_r$ le coefficient d'évolution du but. $g_r$ nous permet d'ajuster la vitesse d'évolution du but et la difficulté de l'atteindre indépendamment du temps ou des caractéristiques techniques de la machine physique sur laquelle sont réalisées les expérimentations.

\subsection{Cadre expérimental}

Étant donné l'évolution aléatoire du but, chaque expérience a été réalisée 100 fois pour un problème et un coefficient d'évolution de but donné. Les résultats présentés par la suite sont donc des moyennes. Toutes les expériences ont été menées sur une machine Linux Fedora 16 équipée d'un processeur Intel Xeon 4 Core (2.0Ghz) avec un maximum de 4Go de mémoire vive et 60 secondes de temps CPU alloués.

Dans un premier temps, les différents algorithmes sont testés sur le domaine Blockworld extrait de IPC-2 afin de mesurer leurs performances respectives en fonction du coefficient d'évolution du but. Dans un deuxième temps, nous présentons quelques résultats montrant l'impact du coefficient de retardement ainsi que de la pondération de l'heuristique (nous présentons ici seulement les résultats obtenus par les meilleurs algorithmes de la première évaluation). Pour terminer, dans un troisième temps, nous proposons une vue d'ensemble des performances des différents algorithmes sur différents domaines et problèmes.

Les performances des algorithmes sont mesurées par: (1) le pourcentage de succès, i.e., le nombre de fois où un algorithme réussit à atteindre le but, (2) le temps mis pour atteindre le but, (3) la longueur des plans solution trouvés.

\subsection{Comparaison des algorithmes sur Blockworld}

\begin{figure}[!]
  \centering
  \subfigure[Pourcentage de succès en fonction du coefficient d'évolution du but.]{\includegraphics[scale=0.9]{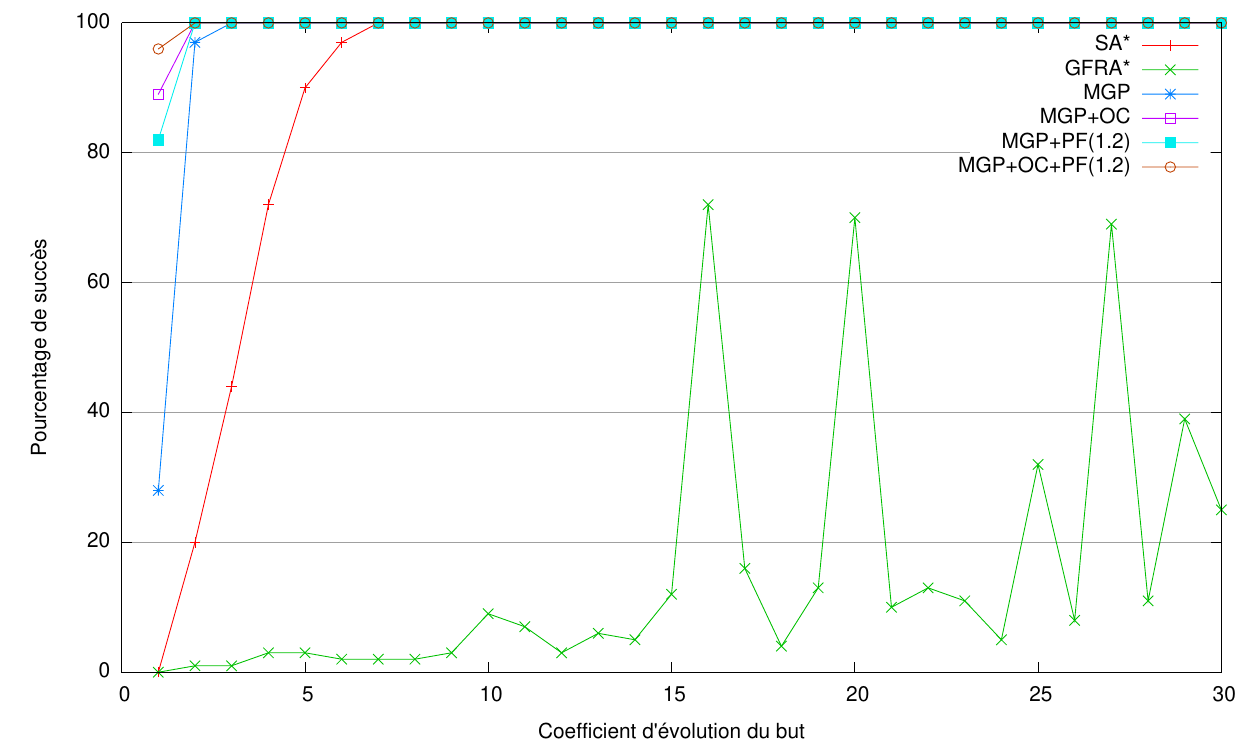} \label{Fig:BW-All-a}}
  \subfigure[Temps de recherche en fonction du coefficient d'évolution du but.]{\includegraphics[scale=0.9]{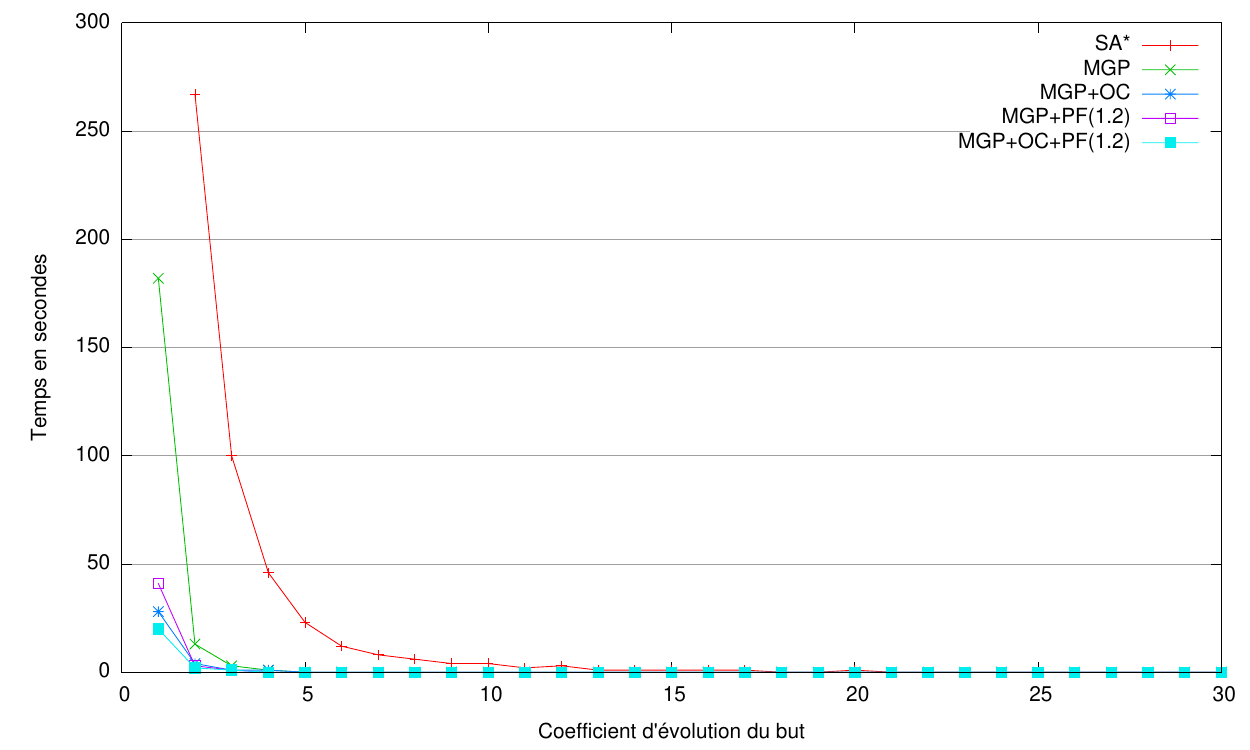} \label{Fig:BW-All-b}}
  \subfigure[Longueurs des plans en fonction du coefficient d'évolution du but.]{\includegraphics[scale=0.9]{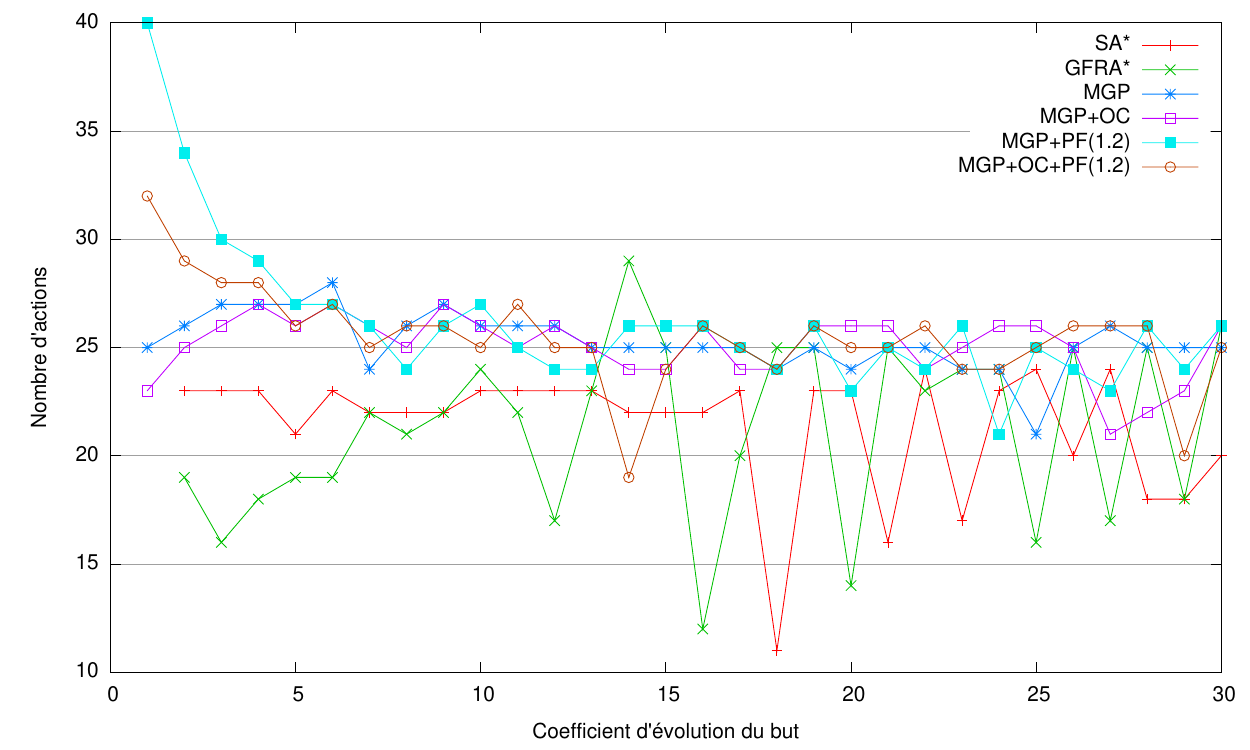} \label{Fig:BW-All-c}}
  \label{Fig:BW-All}
\caption{Analyse comparée des différentes approches sur le problème 20 de Blocksworld.}
\end{figure}

Dans cette partie, nous présentons une comparaison des six algorithmes précédemment introduits -- SA*, GFRA*, MGP, MGP+OC, MGP+PF et MGP+OC+PF -- sur le problème 20 du domaine Blocksworld de la compétition IPC-2 en fonction du coefficient d'évolution du but. Le coefficient de retardement pour MGP+PF et MGP+OC+PF est fixé arbitrairement à 1.2 et le coefficient de pondération de l'heuristique est de 1 pour tous les algorithmes. Les figures~\ref{Fig:BW-All-a}, \ref{Fig:BW-All-b} et \ref{Fig:BW-All-c} présentent respectivement les résultats obtenus en termes de pourcentages de succès, temps de recherche et longueur des plans trouvés.

En ce qui concerne le critère de succès, le meilleur algorithme est MGP+OC+PF. Même avec un coefficient d'évolution du but extrême $g_r = 1$, MGP+OC+PF parvient à trouver une solution dans 95\% des expériences réalisées. Pour les autres stratégies de retardement, les résultats sont légèrement moins bons mais restent très satisfaisants puisqu'ils restent au dessus de 80\% de succès. L'approche naïve SA* nécessite un coefficient d'évolution du but 5 fois plus grand pour obtenir le même pourcentage de succès. GFRA* lui n'atteint pas ce pourcentage de succès même avec un coefficient d'évolution du but 30 fois supérieur. Finalement, GFRA* est largement surclassé par les autres algorithmes.

En terme de temps de recherche, le meilleur algorithme est également MGP+OC+PF. Puis viennent ensuite MGP+OC, MGP+PF et MGP. Finalement, nous avons SA* et très loin GFRA* (GFRA* n'est pas représenté sur la figure~\ref{Fig:BW-All-b}). Une fois encore, MGP+OC+PF surclasse les autres algorithmes. Le couplage des deux stratégies de retardement améliore de manière significative la version naïve de MGP.

S'agissant de la longueur des plans trouvés, MGP et ses différentes variantes produisent des plans plus longs que SA* et GFRA* (presque deux fois plus longs dans le cas extrême où $g_r = 1$). Deux raisons expliquent cette différence. Tout d'abord, rappelons que la stratégie de retardement OC vérifie si le nouveau but est déjà présent dans l'arbre de recherche. Si c'est le cas, MGP extrait alors directement un nouveau plan à partir de l'arbre de recherche. Toutefois, si ce mécanisme limite le nombre d'appel à la procédure de recherche, il ne garantit pas que le plan extrait soit optimal car même si l'heuristique est admissible, l'arbre de recherche n'a pas été construit pour le nouveau but. Deuxièmement, la stratégie de retardement PF (cf. figure~\ref{Fig:BW-All-c}) tend à augmenter la longueur des plans trouvés. Cependant, la différence de qualité des plans trouvés est en grande partie compensée par des temps de recherche et des pourcentages de succès biens meilleurs.

\subsection{Impact du coefficient de retardement et de l'heuristique pondérée}

Dans cette partie, nous évaluons l'impact du coefficient de retardement ainsi que de la pondération de l'heuristique sur le meilleur algorithme obtenu précédemment, i.e., MGP+OC+PF. Les expériences suivantes ont été réalisées sur le même problème (Blockworld problème 20) pour permettre une comparaison des résultats. Étant donné que MGP+OC+PF a un pourcentage de succès proche de 100\% nous ne présentons dans ce qui suit que les résultats en termes de temps de recherche et de longueur de plans.

\paragraph{Impact du coefficient de retardement}

\begin{figure}[!]
  \centering
  \subfigure[Temps de recherche en fonction du coefficient de retardement et du coefficient d'évolution du but.]{\includegraphics[scale=0.9]{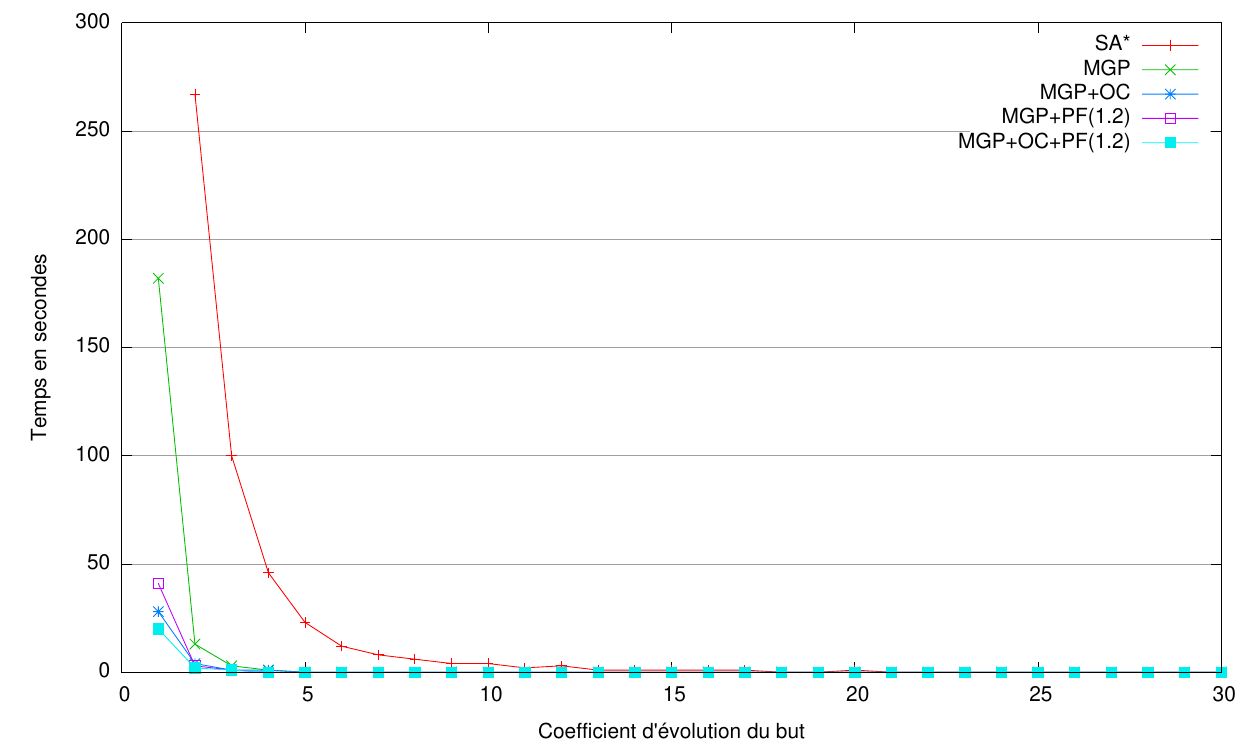} \label{Fig:BW-PF-A}}
  \subfigure[Longueur des plans en fonction du coefficient de retardement et du coefficient d'évolution du but.]{\includegraphics[scale=0.9]{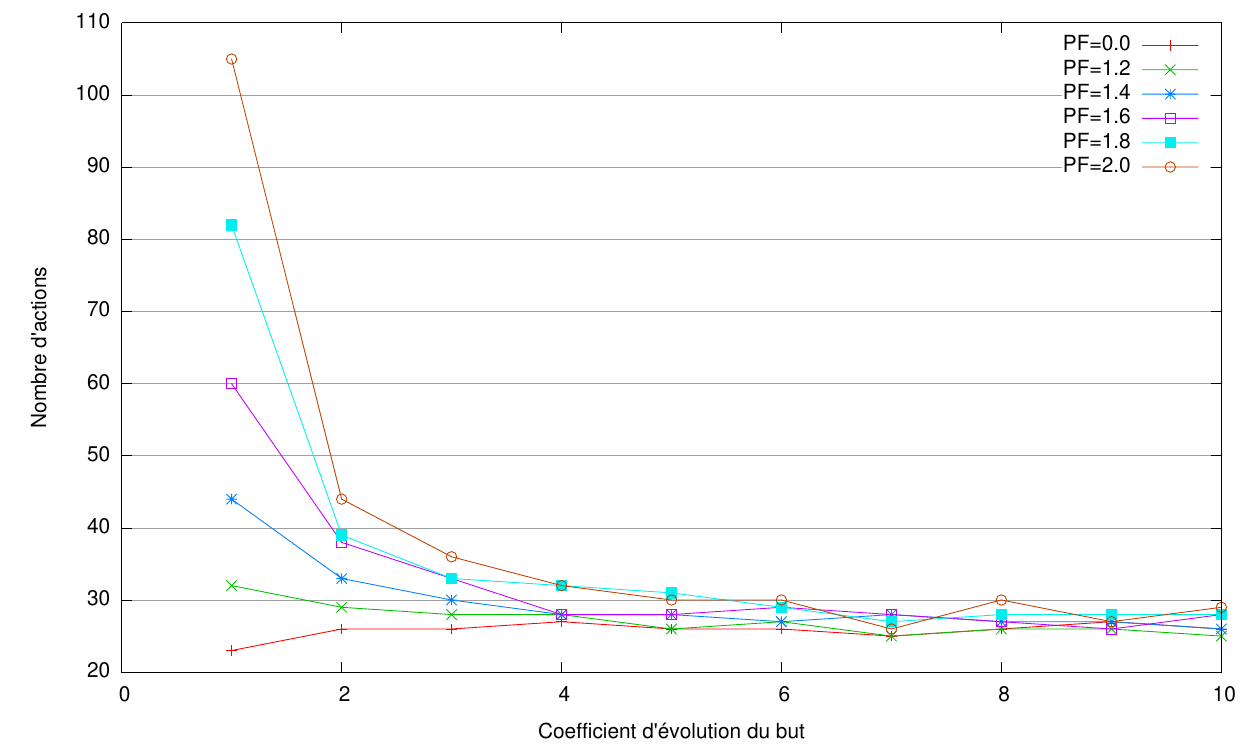} \label{Fig:BW-PF-B}}
  \caption{Impact du coefficient de retardement sur le problème 20 de Blocksworld.}
\end{figure}

Les figures \ref{Fig:BW-PF-A} et \ref{Fig:BW-PF-B} présentent respectivement le temps de recherche et la longueur des plans obtenus en fonction du coefficient d'évolution du but. Nous pouvons faire trois observations. Premièrement, nous pouvons constater que le coefficient de retardement améliore de manière significative les performances de MGP. Par exemple, MGP avec un coefficient de retardement de 2 est 6 fois plus rapide que MGP avec un coefficient de retardement de 0 lorsque l'évolution du but est très rapide ($g_r = 1$). Deuxièmement, on peut constater que le temps de recherche ainsi que la longueur des plans convergent rapidement vers une même valeur et ce quelle que soit par la suite la valeur de $g_r$. Par exemple, avec un coefficient $g_r \geq 4$, le coefficient de retardement n'a plus d'impact sur le temps de recherche et sur la longueur des plans. Troisièmement, on constate qu'une augmentation du coefficient de retardement implique également une augmentation de la longueur des plans trouvés mais réduit le temps de recherche. Par conséquent, le coefficient de retardement doit être choisi de manière à être un compromis entre la longueur des plans trouvés et temps de recherche.

\paragraph{Impact de l'heuristique pondérée}

\begin{figure}[!]
  \centering
  \subfigure[Temps de recherche en fonction de la pondération de l'heuristique et du coefficient d'évolution du but.]{\includegraphics[scale=0.9]{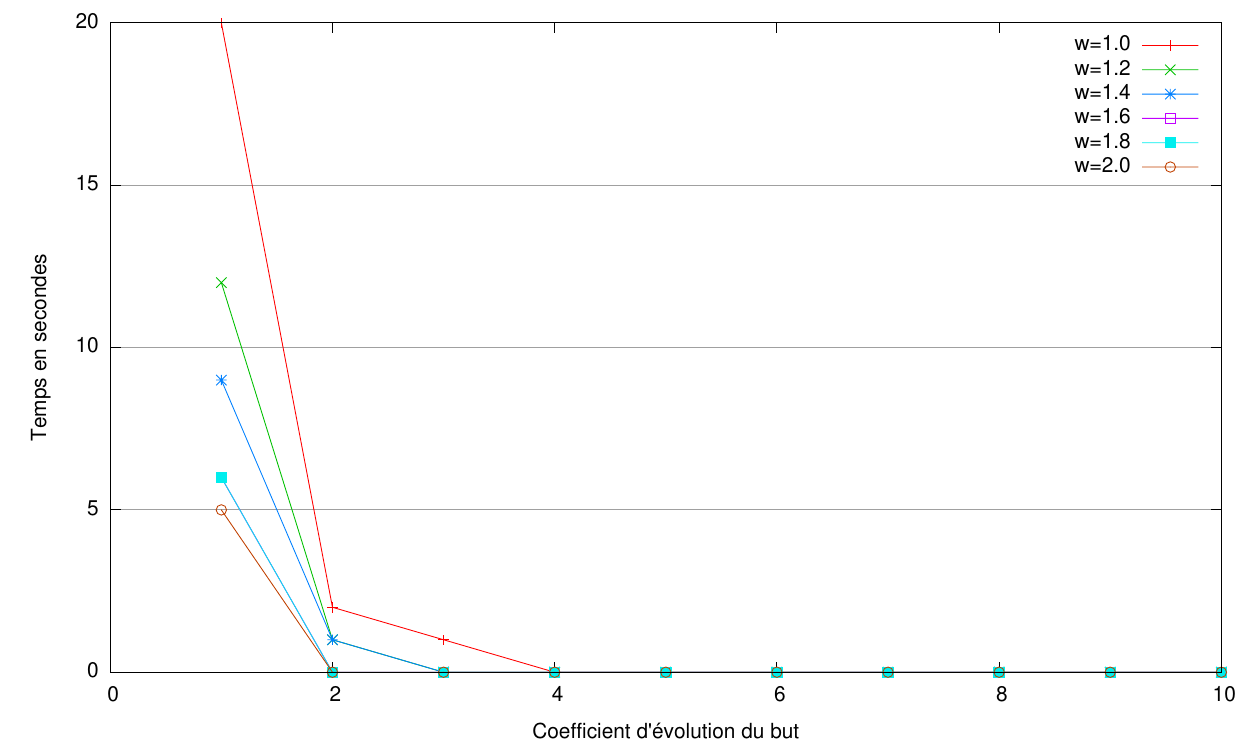} \label{Fig:BW-W-A}}
  \subfigure[Longueur des plans en fonction de la pondération de l'heuristique et du coefficient d'évolution du but.]{\includegraphics[scale=0.9]{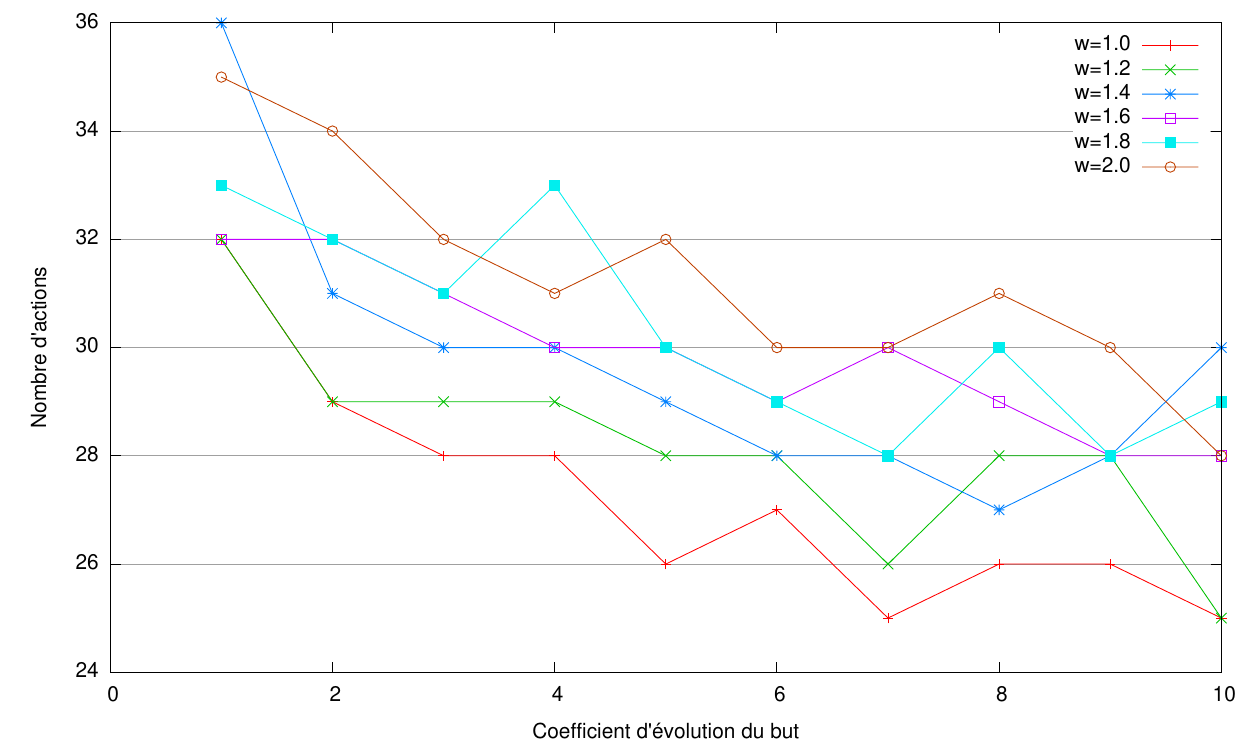} \label{Fig:BW-W-B}}
  \caption{Impact de la pondération de l'heuristique sur le problème 20 de Blocksworld.}
  \label{Fig:Pathfollow}
\end{figure}

Les figures \ref{Fig:BW-W-A} et \ref{Fig:BW-W-B} présentent respectivement le temps de recherche et la longueur des plans obtenus en fonction de la pondération de l'heuristique. La pondération de l'heuristique $w$ améliore également de manière significative les performances de MGP (MGP est 4 fois plus rapide avec $w = 2.0$ qu'avec $w=1.0$ pour un coefficient d'évolution du but  $g_r = 1$). En outre, l'impact de $w$ sur la longueur des plans obtenus n'est pas significatif: quelle que soit la pondération de l'heuristique, la longueur des plans est comparable pour une valeur de coefficient d'évolution de but donné. En d'autres termes, la pondération de l'heuristique augmente les performances de MGP comme on pouvait si attendre, mais ne détériore pas de manière importante la qualité des plans trouvés. Habituellement, l'utilisation d'une heuristique pondérée détériore bien plus la qualité des plans obtenus.

\subsection{Vue d'ensemble des performances sur différents problémes}

Dans cette partie, nous présentons une vue d'ensemble des performances des différents algorithmes sur plusieurs problèmes issus des compétitions IPC: le tableau~\ref{Tab:search_time} donne les temps de recherche obtenus et le tableau~\ref{Tab:success}, les pourcentages de succès. Comme précédemment, chaque expérience a été répétée 100 fois pour obtenir des résultats statistiquement représentatifs. À chaque expérience, un temps CPU de 60 seconds et un maximum de 4Go de mémoire vive ont été alloués. Le coefficient de retardement a été fixé à 1.2.

En terme de temps de recherche, on retrouve les résultats obtenus sur le problème Blockworld 20: MGP+OC+PF est largement plus rapide que les autres algorithmes. De plus, MGP+OC+PF surclasse les variantes de MGP avec une seule stratégie de retardement. De même, MGP+OC+PF surclasse les autres algorithmes en termes de pourcentage de succès. Cependant, on peut constater que MGP sans stratégie de retardement échoue sur certains problèmes (Elevator P35 and Freecell P26) que SA* résout, même si d'une manière générale, la version naïve de MGP reste meilleure que SA* et GFRA*. Finalement, en termes de longueur de plans, toutes les approches trouvent des plans de longueur comparable ce qui ne constitue par conséquent pas un critère discriminant entre les différentes approches.

Pour synthétiser ces résultats, OC et PF améliorent individuellement les performances de MGP qui surclasse SA* et GFRA*. Les deux stratégies de retardement OC+PF couplées donnent de meilleurs résultats que lorsqu'elles sont utilisées séparément. Les résultats obtenus sur différents domaines et problèmes confirment que MGP+OC+PF est le meilleur algorithme et que GFRA* est largement dépassé (GFRA* met à jour la fonction heuristique de tous les états de l'arbre de recherche avant chaque nouvelle recherche ce qui le pénalise comparativement aux autres algorithmes).

\begin{table}
\begin{center}
\begin{tabular}{l||r|r|r|r|r|r}
Problème         & SA*     & GFRA*  & MGP    & OC & PF & OC+PF \\
\hline
depot p03       & 8,12    & -      & 2,40   & 1,28   & 1,67        & 0,78           \\
depot p07       & -       & -      & 7,41   & 6,42   & 1,73        & 1,13           \\
driverlog p03   & 0,03    & 0,02   & 0,33   & 0,22   & 0,18        & 0,17           \\
driverlog p06   & 13,57   & -      & 4,32   & 5,50   & 5,53        & 4,35           \\
elevator p30    & 24,85   & 23,69  & 29,32  & 3,23   & 2,22        & 1,46           \\
elevator p35    & 23,04   & -      & -      & 44,60  & 35,31       & 20,8           \\
freecell p20    & 10,94   & -      & -      & 4,72   & 6,20        & 3,65           \\
freecell p26    & 48,76   & -      & 56,61  & 26,70  & 32,12       & 24,2           \\
pipeworld p04   & 0,50    & 17,73  & 1,68   & 0,49   & 0,55        & 0,48           \\
pipeworld p08   & -       & -      & 18,02  & 13,89  & 13,70       & 12,51          \\
rover p03       & 23,35   & 14,17  & 3,72   & 2,85   & 2,15        & 1,87           \\
rover p07       & -       & -      & -      & 23,51  & -           & 22,54          \\
satellite p03   & -       & 17,26  & 9,36   & 3,67   & 4,56        & 3,18           \\
satellite p06   & -       & -      & -      & -      & -           & 8,97           \\
\end{tabular}
\caption{Comparaison du temps de recherche.}
\label{Tab:search_time}
\end{center}
\end{table}

\begin{table}
\begin{center}
\begin{tabular}{l||r|r|r|r|r|r}
Problème         & SA*    & GFRA* & MGP & OC & PF & OC+PF \\
\hline
depot p03       & 10     & -     & 99  & 30     & 60          & 100            \\
depot p07       & -      & -     & 33  & 12     & 14          & 88             \\
driverlog p03   & 100    & 100   & 100 & 100    & 100         & 100            \\
driverlog p06   & 1      & -     & 1   & 56     & 88          & 98             \\
elevator p30    & 69     & 99    & 91  & 100    & 100         & 100            \\
elevator p35    & 1      & -     & -   & 19     & 5           & 55             \\
freecell p20    & 39     & -     & -   & 99     & 100         & 100            \\
freecell p26    & 56     & -     & 1   & 100    & 100         & 100            \\
pipeworld p04   & 99     & 7     & 99  & 99     & 98          & 100            \\
pipeworld p08   & -      & -     & 48  & 70     & 60          & 76             \\
rover p03       & 48     & 8     & 99  & 99     & 94          & 99             \\
rover p07       & -      & -     & -   & 32     & -           & 52             \\
satellite p03   & -      & 1     & 4   & 16     & 15          & 52             \\
satellite p06   & -      & -     & -   & -      & -           & 28             \\
\end{tabular}
\caption{Comparaison du pourcentage de succès.}
\label{Tab:success}
\end{center}
\end{table}

\begin{table}
\begin{center}
\begin{tabular}{l||r|r|r|r|r|r}
Problème          & SA*     & GFRA*  & MGP   & OC & PF & OC+PF \\
\hline
depot p03       & 20,80   & -      & 21,73 & 21,67  & 25,18       & 22,83           \\
depot p07       & -       & -      & 25,24 & 23,08  & 26,00       & 24,74           \\
driverlog p03   & 7,30    & 4,14   & 8,42  & 11,57  & 12,57       & 9,57            \\
driverlog p06   & 12,00   & -      & 14,00 & 15,00  & 11,23       & 16,91           \\
elevator p30    & 29,20   & 27,88  & 27,33 & 28,42  & 27,74       & 28,80           \\
elevator p35    & 34,00   & -      & -     & 33,05  & 32,20       & 32,18           \\
freecell p20    & 29,97   & -      & -     & 29,99  & 29,99       & 29,99           \\
freecell p26    & 37,02   & -      & 38,00 & 37,01  & 37,01       & 37,01           \\
pipeworld p04   & 3,21    & 11,86  & 7,61  & 9,19   & 10,20       & 8,12            \\
pipeworld p08   & -       & -      & 18,21 & 21,09  & 19,76       & 21,02           \\
rover p03       & 44,53   & 44,73  & 44,40 & 44,54  & 44,66       & 44,24           \\
rover p07       & -       & -      & -     & 43,20  & -           & 43,50           \\
satellite p03   & -       & 42,00  & 26,00 & 24,69  & 32,12       & 25,80           \\
satellite p06   & -       & -      & -     & -      & -           & 26,00           \\
\end{tabular}
\caption{Comparaison de la longueur des plans obtenus.}
\label{Tab:plan_length}
\end{center}
\end{table}

\section{Conclusion}

Dans cet article, nous avons proposé une approche pour la planification de tâches en boucle fermée, appelée MGP, qui adapte continuellement son plan aux évolutions successives du but. MGP s'appuie sur une recherche heuristique pondérée incrémentale de type A* et entrelace planification et exécution. Afin de limiter le nombre de recherches effectuées pour prendre en compte les changements dynamiques du but au cours du temps, MGP retarde autant que possible le déclenchement de nouvelles recherches. Pour cela, MGP utilise deux stratégie de retardement: {\it Open Check} (OC) qui vérifie si le but est encore présent dans l'arbre de recherche et {\it Plan Follow} (PF) qui estime s'il est préférable d'exécuter les actions du plan courant pour se rapprocher du nouveau but plutôt que de relancer une nouvelle recherche. De plus, MGP utilise une stratégie conservatrice et efficace de mise à jour incrémentale de l'arbre de recherche lui permettant de réduire le nombre d'appels à la fonction heuristique et d'accélérer la recherche d'un plan solution.

Nous avons montré expérimentalement que MGP surclassait l'approche naïve SA* et l'algorithme de référence GFRA*. Nous avons également montré que: (1) la combinaison des deux stratégies de retardement OC+PF donnait de meilleurs résultats que chacune considérée individuellement; (2) le coefficient de retardement qui permet de paramétrer la stratégie PF doit être un compromis entre la longueur des plans trouvés et le temps de recherche et (3) la pondération de l'heuristique améliore de manière significative les performances de MGP sans trop détériorer la qualité des plans trouvés.

\end{document}